\documentclass{amia}
\usepackage{lipsum} %Remove if not needed
\usepackage{booktabs} % To thicken table lines
\usepackage{multirow} 
\usepackage{xspace}
\usepackage{smile}
\usepackage{amsfonts}
\usepackage{hyperref}
\usepackage{url}
\usepackage{xcolor,colortbl}
\usepackage{listings}

\newcommand{\dataset}{{CRADLE}\xspace}
\newcommand{\ours}{{EHR-CoAgent}\xspace}

\lstdefinestyle{mystyle}{
  basicstyle=\ttfamily,
  frame=single,
  breaklines=true,
  breakindent=0pt,
  backgroundcolor=\color{gray!10}, % Change the background color
}

\usepackage[superscript,biblabel]{cite}

\usepackage{xcolor,colortbl}
\usepackage{adjustbox}
\usepackage{subfig}
 \usepackage{enumitem}
 
\newif\ifsubmit
\submitfalse
\definecolor{darkpink}{rgb}{0.91, 0.33, 0.5}

\ifsubmit

\else

\fi

\begin{document}

% \title{A Synergistic Approach to Few-Shot EHR Prediction: Combining Predictive LLM Agent Reasoning and Critical LLM Agent Instruction}
\title{LLMs-based Few-Shot Disease Predictions using EHR: A Novel Approach Combining Predictive Agent Reasoning and Critical Agent Instruction}
% (Combined with a Critical Agent) 
\author{Hejie Cui$^1$, Zhuocheng Shen$^1$, Jieyu Zhang$^2$, Hui Shao, MD, PhD$^{4,5}$, Lianhui Qin, PhD$^3$, Joyce C. Ho, PhD$^1$, Carl Yang, PhD$^{1,4}$}

\institutes{
    $^1$ Department of Computer Science, Emory University, Atlanta, GA  \\
    $^2$ School of Computer Science \& Engineering, University of Washington, Seattle, WA \\
    $^3$ Department of Computer Science \& Engineering, UCSD, San Diego, CA \\
    $^4$ Rollins School of Public Health, Emory University, Atlanta, GA\\
    $^5$ School of Medicine, Emory University, Atlanta, GA
}

\maketitle

\section*{Abstract}
\vspace{-1em}
Electronic health records (EHRs) contain valuable patient data for health-related prediction tasks, such as disease prediction. Traditional approaches rely on supervised learning methods that require large labeled datasets, which can be expensive and challenging to obtain. In this study, we investigate the feasibility of applying Large Language Models (LLMs) to convert structured patient visit data (e.g., diagnoses, labs, prescriptions) into natural language narratives. We evaluate the zero-shot and few-shot performance of LLMs using various EHR-prediction-oriented prompting strategies. Furthermore, we propose a novel approach that utilizes LLM agents with different roles: a predictor agent that makes predictions and generates reasoning processes and a critic agent that analyzes incorrect predictions and provides guidance for improving the reasoning of the predictor agent. Our results demonstrate that with the proposed approach, LLMs can achieve decent few-shot performance compared to traditional supervised learning methods in EHR-based disease predictions, suggesting its potential for health-oriented applications.

\section*{Introduction}
\vspace{-1em}
Large Language Models (LLMs) have emerged as a powerful tool in various domains, including healthcare. These models, such as GPT family~\cite{achiam2023gpt} and PaLM~\cite{anil2023palm}, are trained on vast amounts of text data, allowing them to encode extensive knowledge across multiple fields. 
% Moreover, recent explorations have extended beyond single-LLM agents, demonstrating the potential of LLMs to collaborate in various tasks~\cite{wu2023autogen}. 
In the medical domain, the ability of LLMs to leverage their encoded medical knowledge has been showcased in recent studies~\cite{singhal2023large, hernandez2023we}, with impressive performance on tasks such as medical question answering~\cite{singhal2023towards}, clinical text summarization~\cite{van2024adapted}, and clinical decision support~\cite{hegselmann2023tabllm}. 
% The inherent knowledge possessed by certain very large language models makes them particularly well-suited for few-shot learning, as they can draw upon their existing understanding to quickly adapt to new tasks with limited examples~\cite{brown2020language, schick2020exploiting}. 
Certain very large language models demonstrate an emerging ability for few-shot learning, where the model can draw upon their existing understanding to quickly adapt to new tasks with limited examples~\cite{brown2020language, schick2020exploiting}. 
This raises the question of whether LLMs can be directly applied to perform few-shot disease predictions using Electronic Health Record (EHR) data.

EHRs contain a wealth of patient data for predictive modeling tasks such as disease prediction, readmission risk assessment, and mortality prediction~\cite{shickel2017deep}. Existing approaches to EHR-based prediction primarily rely on supervised learning methods, including traditional machine learning models, representation learning~\cite{rajkomar2018scalable,landi2020deep, attn-modeling-ehr}, and graph-based models~\cite{choi2020learning}. While effective, these supervised approaches require training on large labeled datasets, which can be computationally expensive and challenging to obtain due to the high cost and difficulty of acquiring high-quality labeled EHR data~\cite{xiao2018opportunities}.
%In contrast, few-shot learning methods offer a promising alternative by enabling models to learn from a limited number of demonstrations~\cite{ravi2016optimization,brown2020language}. By reducing the need for extensive labeled training data, few-shot learning techniques can significantly improve the efficiency and feasibility of developing predictive models for EHR data~\cite{wornow2024ehrshot}. Applying LLMs to few-shot disease prediction using EHR data can potentially bring benefits such as reducing the need for large labeled datasets, enabling more rapid development and deployment of predictive models, and facilitating personalized medicine by adapting to individual patient characteristics with just a few examples.
In contrast, the capacity for few-shot learning enables LLMs to adapt to new tasks with minimal data, without any finetuning~\cite{brown2020language}. This adaptability raises the possibility of employing LLMs for few-shot disease prediction using EHR, a step forward in making healthcare more precise and efficient~\cite{wornow2024ehrshot}.

In this study, we investigate the efficacy of LLMs-based few-shot disease prediction using the EHRs generated from clinical encounters that include three types of medical codes: disease, medications, and procedures. We convert the structured patient visit records into unstructured language narratives by mapping the ICD codes to their names and connecting them with proper conjunctives. This conversion process allows LLMs to better understand clinical records and retrieve related internal knowledge. We assess the zero-shot and few-shot diagnostic performance of LLMs using various prompting strategies, such as considering factor interactions and providing prevalence statistics and exemplars. The results of this evaluation provide insights into the potential of LLMs as a tool for EHR-based disease prediction and highlight the influence of prompting strategies on their performance.

\begin{figure}[t]
\centering
  \includegraphics[width=\linewidth]{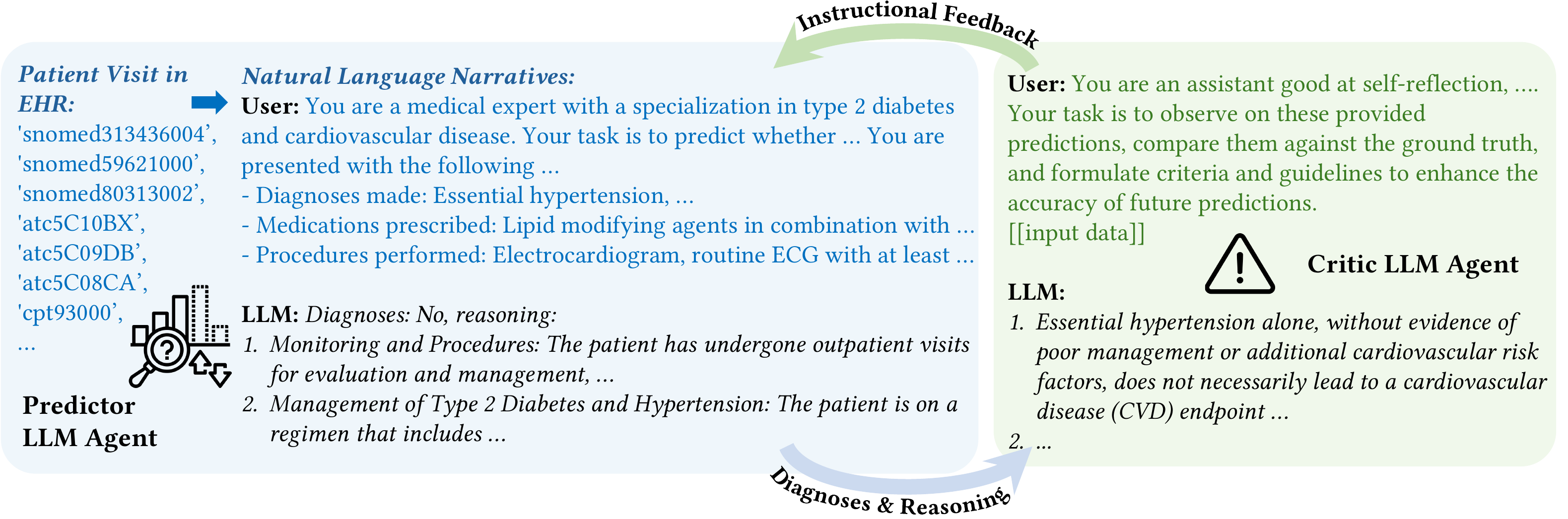}%
  \vspace{-2ex}
    \caption{The framework of {\ours} employs two LLM agents: a predictor agent that makes predictions and generates reasoning processes and a critic agent that analyzes incorrect predictions and provides guidance for improvement. The critic agent’s feedback is used to update the prompts given to the predictor agent, enabling the system to learn from its mistakes and adapt to the specific challenges of the EHR-based disease prediction task.}
    \vspace{-2ex}
    \label{fig:framework}
    % \vspace{-1em}
\end{figure}

Building upon the findings of our initial evaluation, we propose an innovative approach to further improve the few-shot diagnostic performance of LLMs on EHR data. Studies have shown the promise of specialized LLM agents working collaboratively \cite{wu2023autogen,talebirad2023multi,jin2024genegpt}, leveraging their diverse functionalities through few-shot learning.
Our approach combines the strengths of predictive agent reasoning and critical agent instruction to create a more robust and accurate prediction system. The overall framework is shown in Figure~\ref{fig:framework}. Specifically, we employ two LLM agents with different roles: a \textit{predictor agent} and a \textit{critic agent}. The \textit{predictor agent} makes few-shot predictions given the unstructured narratives, which are converted from structured records, and generates a reasoning process to support its predictions. The \textit{critic agent} then takes the predictor's output alongside the ground-truth disease labels as input and identifies issues or biases in the predictor agent's reasoning process. Based on the analysis, the critic agent generates a set of instructions that draw the predictor agent's attention to potentially overlooked factors and offer specific recommendations for refining its reasoning process. These instructions are subsequently appended to the prompts used for the predictor agent, serving as additional context to inform its predictions. Our results show that by refining the prompts based on the critic agent's feedback, the overall diagnostic accuracy of the LLM-based few-shot prediction system improves significantly. This approach leverages the complementary strengths of predictive reasoning and critical analysis, enabling the system to learn from its mistakes and adapt to the specific challenges of EHR-based disease prediction.
In summary, our main contributions are:
\begin{itemize}[nosep,leftmargin=*]

\item We investigate the application of LLMs to EHR-based disease prediction tasks by converting structured data into natural language narratives and evaluating zero-shot and few-shot performance using various prompting strategies.

\item We propose a novel approach combining two LLM agents with different roles: a predictor agent that makes predictions and provides reasoning processes, and a critic agent that analyzes incorrect predictions and provides feedback for improvement. The critic agent's feedback is used to update the predictor agent's prompts, enabling the system to learn from its mistakes and adapt to EHR-based disease prediction challenges.

\item We summarize a set of insights into the performance of LLMs under various settings and share practical guidance on leveraging LLMs for diagnostic tasks with limited labeled data. We hope this can contribute to developing efficient and effective clinical decision support systems in the era of LLMs.
\end{itemize}

\section*{Related Work}
\vspace{-1em}
\underline{\textit{Large Language Models for Healthcare}}
LLMs have demonstrated remarkable capabilities in various application scenarios. Recently, there has been a growing interest in applying LLMs to the medical domain~\cite{thirunavukarasu2023large,he2023survey,cpeng23}, particularly for tasks such as clinical note analysis~\cite{agrawal-etal-2022-large, mannhardt2024impact}, medical question answering~\cite{LIEVIN2024100943, han2023medalpaca}, disease prediction~\cite{wang2023clinicalgpt}, clinical trial matching~\cite{yuan2023large}, medical report generation~\cite{d2024large}. 
For example, Yang et al.~\cite{GatorTron} introduced GatorTron, an LLM specifically designed for EHRs. They demonstrated the effectiveness of GatorTron in various clinical natural language processing (NLP) tasks, such as named entity recognition and relation extraction, showcasing the potential of LLMs to extract valuable information from unstructured EHR data. Peng et al.~\cite{cpeng23} investigated the use of generative LLMs for medical research and healthcare. They explored the capabilities of LLMs in tasks such as medical question answering, disease prediction, and clinical trial matching, highlighting their potential to support clinical decision-making and assist research.

However, applying LLMs to EHR-based disease prediction tasks remains under-explored. While some studies have investigated the use of LLMs for clinical NLP tasks on EHR~\cite{GatorTron}, there is still a lack of research on leveraging the reasoning and instruction-following capabilities of LLMs for few-shot EHR-based prediction. Our research addresses this gap by exploring the use of LLMs for EHR-based disease prediction and proposes new methods to enable accurate prediction with minimal training data.

\section*{Method}
\vspace{-1em}
In this study, we expand our investigations on two levels: (1) evaluating the zero-shot and few-shot performance of LLMs on EHR-based disease prediction tasks, and (2) proposing a novel approach that leverages collaborative LLM agents to enhance the predictive performance. 

\underline{\textit{LLM Performance on Disease Prediction with EHR}}
The structured patient visit data are typically stored in tabular formats, where each row represents an individual patient visit record generated from clinical encounters, and columns correspond to different medical codes. In this study, we utilize EHR data that includes three types of medical codes $\mathcal{C}$: (1) diseases $\mathcal{C}_D$, (2) medications $\mathcal{C}_M$, and (3) procedures $\mathcal{C}_P$. Each patient visit sample $v_i$ in the record $\mathcal{V}$ is represented by a set of medical codes $\{c_1, c_2, \ldots, c_n\}$, where $c_j \in \mathcal{C}$. We convert the structured EHR records into unstructured language narratives, denoted as $\mathcal{H}$, by mapping the medical codes to their names to enable the application of LLMs. 

\noindent $\diamond$  \textbf{\small{Zero-Shot: Leveraging Pre-existing Knowledge}} 
\label{ssec:zero-shot}
% To enhance the zero-shot performance of LLMs on EHR-based prediction tasks and enable improved predictions without task-specific training data, we incorporate various prompting strategies to leverage the models' language understanding capabilities and medical knowledge. 
Prompt engineering has emerged as a powerful technique for guiding the behavior of LLMs and improving their performance on various healthcare-related tasks, such as clinical named entity recognition~\cite{sivarajkumar2023empirical, 10.1093/jamia/ocad259} and clinical text classification~\cite{lu2023medical,sivarajkumar2022healthprompt}. We develop a set of prompting strategies tailored to EHR-based prediction tasks to provide additional context and guide the reasoning process of LLMs, including: 
\begin{itemize}[nosep,leftmargin=1em]
\item Chain-of-thought (CoT) reasoning~\cite{wei2022chain}: prompt the LLMs to generate step-by-step explanations;
\item Incorporation of factor interactions: encourage LLMs to consider the interactions and dependencies among different medical factors (e.g., diseases, medications, and procedures);
\item Prevalence information: integrate information about the prevalence statistics to provide additional context.
\end{itemize}

\noindent $\diamond$  \textbf{\small{Few-Shot: Enhancing Performance with Limited Examples}} 
We randomly select a small number of positive and negative samples (e.g., 3 positive and 3 negative) from the training data to serve as exemplars for each prediction category. These exemplars are incorporated into the prompts to provide the LLMs with a limited set of task-specific examples to learn from. 
This leverages the LLMs' vast pre-existing knowledge while allowing them to adapt quickly to the specific characteristics of the EHR prediction task. By this, we aim to guide LLMs' attention toward the most relevant patterns associated with each prediction category. 

\underline{\textit{\ours: Collaborative LLM Agents for Enhanced Prediction}}
Recently, the potential of LLMs has extended beyond single-agent applications. By leveraging the power of multiple LLMs with different roles working together in a collaborative framework, new possibilities have been unlocked for tackling complex problems and enhancing the performance of language models~\cite{wu2023autogen}.
In this study, we propose a novel approach called \ours (as demonstrated in Figure~\ref{fig:framework}), which harnesses the potential of collaborative LLM agents for enhanced prediction of EHR. Our framework consists of two components: a predictor agent $\mathcal{P}_{\text{LLM}}$ and a critic agent $\mathcal{K}_{\text{LLM}}$. The predictor agent focuses on generating predictions and providing explanatory reasoning, while the critic agent observes the predictor's outputs and provides instructional feedback to refine the prediction process. By integrating the feedback from the critic agent into the prompts used by the predictor agent, we aim to create an in-context learning process with feedback to continuously enhance disease prediction accuracy.

\begin{figure}[ht]
\centering
  \includegraphics[width=\linewidth]{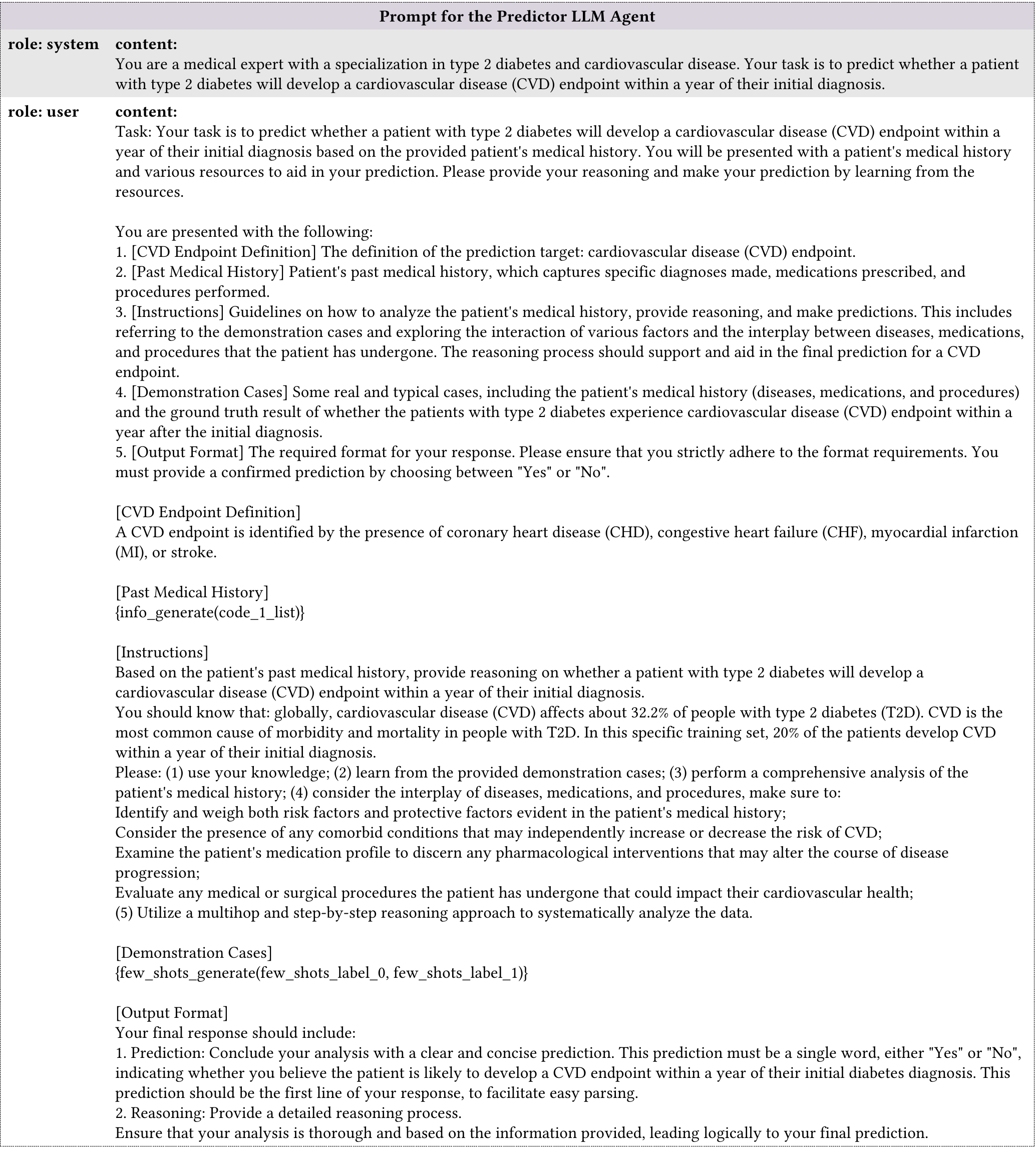}%
    \caption{Prompt for Predictor Agent in {\ours} for the CRADLE dataset.}
    \label{fig:prompt_predictor}
    \vspace{-1em}
\end{figure}

\noindent $\diamond$  \textbf{\small{Predictor Agent: Generating Predictions and Reasoning}} 
The predictor agent $\mathcal{P}_{\text{LLM}}$ is an LLM that performs few-shot disease predictions and provides explanatory reasoning based on the input EHR data. Given a patient's medical history $\mathcal{H}_i$, the predictor LLM analyzes the relevant information and generates the most likely prediction $\hat{\mathcal{D}_i}$ and provides a step-by-step explanation of its reasoning process $\mathcal{R}_i$. Such explanatory reasoning is crucial for enhancing the interpretability of the generated predictions. By highlighting the key factors and evidence influencing the LLM agent's decision-making process, the reasoning serves as a transparent and informative basis for further analysis and validation. The detailed prompt we used for the predictor agent in \ours is shown in Figure~\ref{fig:prompt_predictor}.

\begin{figure}[ht]
\centering
  \includegraphics[width=\linewidth]{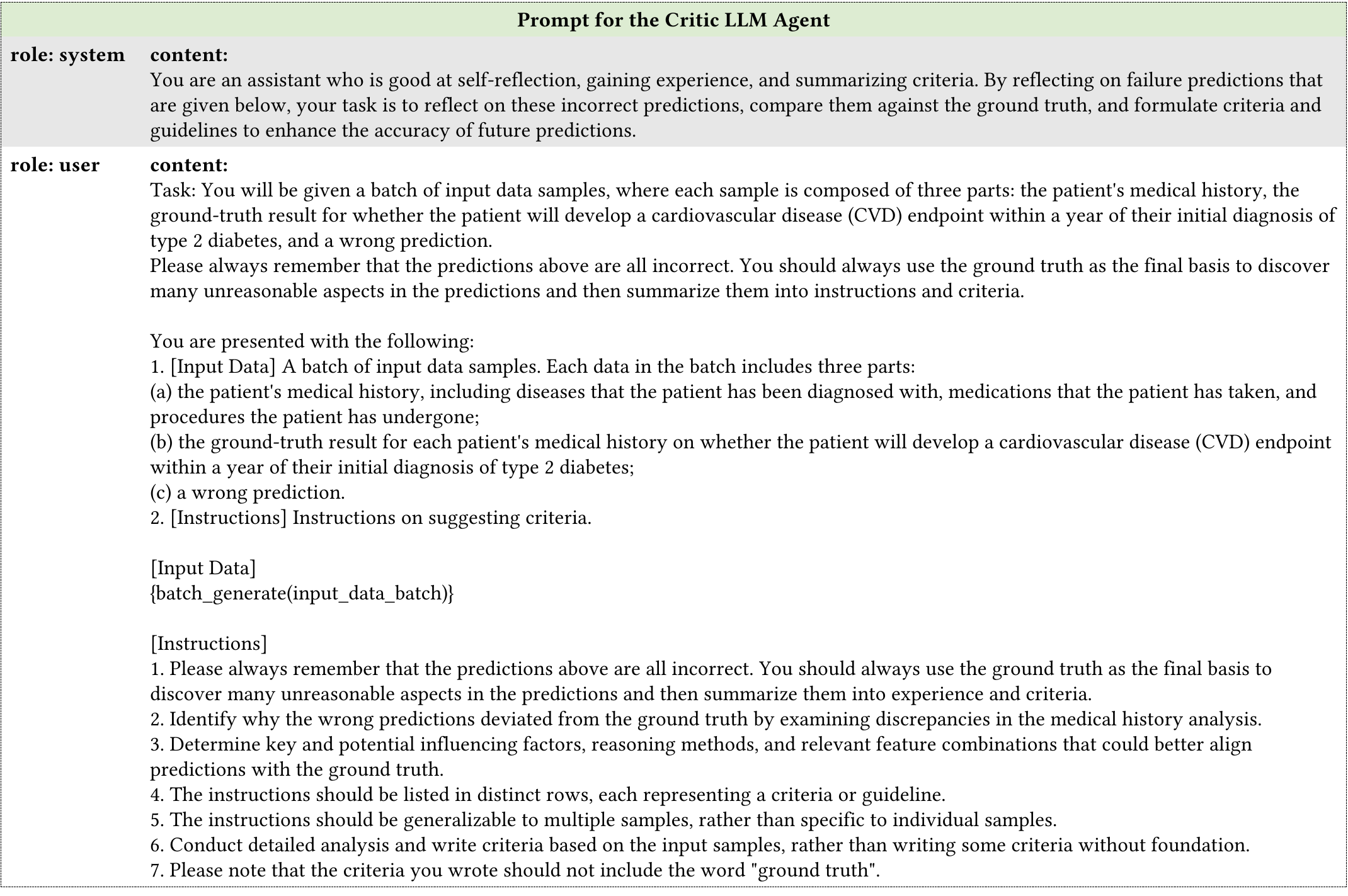}%
    \caption{Prompt for Critic Agent in {\ours} for the CRADLE dataset.}
    \label{fig:prompt_critic}
    \vspace{-1em}
\end{figure}

\noindent $\diamond$  \textbf{\small{Critic Agent: Providing Instructional Feedback}} 
The critic agent $\mathcal{K}_{\text{agent}}$ is another LLM that plays a different role in the \ours framework by observing a set of sampled wrong predictions from the predictor agent. Each set, denoted as $\mathcal{B}_j = \{(\hat{\mathcal{D}}_{ji}, \mathcal{R}_{ji})\}_{i=1}^b$, contains generated prediction $\hat{\mathcal{D}}_{ji}$ and their accompanying explanatory reasoning $\mathcal{R}_{ji}$ for $b$ instances. The critic agent analyzes the inconsistency of the generated prediction to their corresponding ground truth label $\mathcal{D}_{ji}$ for each batch $\mathcal{B}_j$, identifying error patterns for improvement. Based on this analysis, we let the critic agent generate a set of instructional feedback $\{\mathcal{F}_j\}$ for batch $\mathcal{B}_j$ and repeat this process for $m$ times. The detailed prompt we used for the critic agent in \ours is shown in Figure~\ref{fig:prompt_critic}.

To provide concise and coherent guidance, we employ GPT-4 to process the set of instructional feedback $\{\mathcal{F}_j\}_{j=1}^m$. GPT-4 analyzes the feedback across multiple batches and generates a consolidated set of instructions $\mathcal{F}_{\text{consolidated}}$ that captures the most important and recurring insights. This consolidated feedback highlights common biases or errors in the reasoning process, offers suggestions for considering additional factors,  and provides insights into the relationships between different medical concepts. 

\noindent $\diamond$  \textbf{\small{Instruction-Enhanced Prompting: Integrating Feedback for Refinement}} 
To effectively incorporate the feedback generated by the critic LLM, we introduce an instruction-enhanced prompting mechanism. This mechanism integrates the critic LLM's instructional feedback $\mathcal{F}_{\text{consolidated}}$ directly into the prompts $\mathcal{P}$ used by the predictor LLM. By augmenting the prompts with specific instructions and guidance, we aim to steer the predictor LLM's attention toward the most relevant aspects of the input data and encourage it to consider the insights provided by the critic LLM. This iterative process of making predictions, receiving feedback, and refining the prompts allows the predictor LLM to continuously improve its performance and adapt to the specific challenges of EHR-based disease prediction.

\section*{Experimental Settings}
\vspace{-1em}
\underline{\textit{Datasets}}
We conducted experiments on two datasets: the publicly accessible MIMIC-III dataset and the privately-owned CRADLE dataset.
\textbf{MIMIC-III}~\cite{johnson2016mimic} is a large, publicly accessible dataset comprising de-identified health-related data associated with over forty thousand patients who stayed in critical care units of the Beth Israel Deaconess Medical Center between 2001 and 2012. Our task is to predict whether acute care conditions will be present during a patient's next visit, given their current ICU stay records. We focus on a specific chronic phenotype, Disorders of Lipid Metabolism, which is identified using Clinical Classifications Software (CCS) from the Healthcare Cost and Utilization Project (HCUP)\footnote{\url{https://www.hcup-us.ahrq.gov/toolssoftware/ccs/AppendixASingleDX.txt}}. 
During preprocessing, we extract patients with more than one hospital visit and create pairs of adjacent visits for each patient. For each pair, the former visit serves as the input, and the phenotypes in the latter visit are used as labels. This process yields 12,353 records with labels. For budget consideration, we randomly sample 1,000 records based on the data distribution of the prediction target as our testing set.

Project \textbf{CRADLE} (Emory Clinical Research Analytics Data Lake Environment) is a privately-owned database that contains de-identified electronic health records at Emory Healthcare from 2013 to 2017. In this study, we focus on the patients with type 2 diabetes and predict whether those patients will experience \textbf{cardiovascular disease} (CVD) endpoints within a year after the initial diabetes diagnosis. The CVD endpoints include coronary heart disease (CHD), congestive heart failure (CHF), myocardial infarction (MI), or stroke, which are identified by their ICD-9 and ICD-10 clinical codes. For patients who developed CVD complications within a year (positive cases), we select the earliest recorded encounter within a year of the CVD endpoint presence as the input. For patients without CVD complications (negative cases), we randomly select one encounter as the input from all encounters that occurred at least one year before the last recorded encounter. Patients are excluded if they (1) have less than two encounters at Emory Healthcare, (2) the time interval between their first and last encounter is less than one year, or (3) have a history of CVD conditions. After applying these exclusion criteria, 35,404 patients remain in the dataset. Similar to MIMIC-III, we randomly sample 1,000 records based on the data distribution of the prediction target

% The sample characteristics of the two datasets are presented in existing works~\cite{johnson2016mimic,ho2023evaluation}, and detailed statistics are shown in Table~\ref{tab:dataset}.

% \begin{table}[h]
%     \centering
%     \renewcommand\arraystretch{0.9}
% % \resizebox{\linewidth}{!}{
% % \begin{minipage}[p][0.5\textwidth]
% \begin{tabular}{lcc}
% \toprule
% \bfseries Stats & \bfseries MIMIC-III    & \bfseries \dataset          \\ 
% \midrule 
% \# of diagnosis & 846 & 7915\\
% \# of medication & 4525 & 489\\
% \# of procedure & 2032 & 4321\\
% \# of service & 20 & ---\\
% \# of hyperedges & 36875/12353 & 36611 \\
% Prevalence & 27.6\% & 21.4\%\\
% \bottomrule
% \end{tabular}
% % \vspace{-1ex}
% % }
% %  \end{minipage}
%  \caption{Dataset statistics. For \# of hyperedges in MIMIC-III, the first number indicates the hyperedges without labels, while the second one indicates ones with labels.}
% \vspace{-1ex}
% \label{tab:dataset}
% \end{table}

\underline{\textit{Evaluation Metrics}}
Both the MIMIC-III and CRADLE datasets exhibit class imbalance, with the prevalence of Disorders of Lipid Metabolism in MIMIC-III being 27.6\% and the prevalence of cardiovascular disease (CVD) endpoints in CRADLE being 21.4\%. To account for the imbalanced data distributions, we employ accuracy, sensitivity, specificity, and F1 score as evaluation metrics~\cite{choi2020learning}. 
%For machine learning methods, a threshold of 0.5 is used to classify the predicted scores into their respective classes. 
When evaluating LLM methods, we identify the presence of ``Yes" or ``No" tokens in the LLM responses and extract the top 5 probabilities associated with the predicting token. These probabilities are then normalized over both answers. We observed that GPT family models tend to provide highly confident answers (a confirmed prediction of either ``Yes'' or ``No'', with almost 0.0 probability for the other choice), often resulting in a majority probability of either 0.0 or 1.0. 
% Consequently, probability-based metrics such as Area Under the Receiver Operating Characteristic curve (AUROC) and Area Under the Precision-Recall curve (AUPR) are not suitable for calculation in this context, as they may not accurately reflect the performance of the LLM methods given the highly skewed probability distributions.

\underline{\textit{Baselines}}
We compare the performance of \ours with traditional machine learning (ML), including Decision Trees, Logistic Regression, and Random Forests, which are widely used in EHR-based prediction tasks~\cite{wu2010prediction,goldstein2017opportunities}, and single-agent LLM approaches using GPT-4 (\texttt{gpt-4-0125-preview}) and GPT-3.5 (\texttt{gpt-35-turbo-16k-0613}). 
The ML models are trained in both fully supervised and few-shot settings, while the LLM approaches are evaluated in pure zero-shot, zero-shot with additional prompt information as mentioned in section~\ref{ssec:zero-shot}, and few-shot learning settings. 
%The zero-shot setting relies solely on the LLM's pre-existing knowledge, while Enhanced Zero-Shot incorporates additional prompting techniques to improve performance. The few-shot setting provides the LLM with a small number of labeled examples to learn from. 
By comparing \ours with these baselines, we aim to evaluate the effectiveness of diverse LLM agent frameworks in EHR-based disease prediction tasks.

% [1] Rajkomar, A., Oren, E., Chen, K., Dai, A. M., Hajaj, N., Hardt, M., ... & Dean, J. (2018). Scalable and accurate deep learning with electronic health records. NPJ Digital Medicine, 1(1), 1-10.

% [2] Huang, K., Altosaar, J., & Ranganath, R. (2019). ClinicalBERT: Modeling clinical notes and predicting hospital readmission. arXiv preprint arXiv:1904.05342.

\underline{\textit{Implementation Details}}
We implemented the empirical study methods in Python. The baseline machine learning models were trained and evaluated using the popular sklearn package, which provides a comprehensive set of tools for machine learning tasks. To access the various GPT models securely, we utilized the Azure OpenAI Service, a trusted and compliant cloud platform. Azure OpenAI offers a secure API interface that allows seamless integration of the GPT capabilities into our research pipeline while maintaining strict privacy and security controls. By leveraging Azure OpenAI, we ensured that the sensitive patient dataset was processed in a protected environment, adhering to necessary regulations and standards, such as HIPAA and GDPR.     
%such as HIPAA (Health Insurance Portability and Accountability Act) and GDPR (General Data Protection Regulation). 
% This approach guarantees that the data remains confidential and is not exposed to unauthorized access, providing a reliable and secure framework for our research. 

\section*{Experimental Results}
% prediction bias of zero-shot 
% prompt design 
% insights: class 0 
\vspace{-1em}
\begin{table*}[h]
% \setlength{\tabcolsep}{1pt}
% \resizebox{\linewidth}{!}{
%   \begin{adjustbox}{width=\columnwidth,center}
%   \resizebox{\columnwidth}{!}{
\vspace{-1ex}
\renewcommand\arraystretch{0.9}
% \vspace{-12pt}
\begin{center}
\caption{Performance (\%) of different models under the zero-shot, few-shot, and fully-supervised settings on MIMIC-III and \dataset datasets. The proposed method is colored in \colorbox{green!10}{green}. The reference results under the supervised training setting (trained on 11,353 samples for MIMIC-III and 34,404 samples for CRADLE) are colored in \colorbox{gray!10}{gray}.}
\vspace{-1em}
\resizebox{\linewidth}{!}{
\begin{tabular}{llccccccccc}
\toprule
\multirow{2.5}{*}{\bfseries Type} & \multirow{2.5}{*}{\bfseries Model} & \multicolumn{4}{c}{\bfseries MIMIC-III (Pos : Neg =  27.6\% : 72.4\%)}  &  &  \multicolumn{4}{c}{\bfseries \dataset (Pos : Neg =  21.4\% : 78.6\%)}           \\ 
\cmidrule(lr){3-6} \cmidrule(lr){8-11}
& &  ACC &  Sensitivity &  Specificity  & F1 & & ACC &  Sensitivity & Specificity &  F1\\
\midrule
% \multicolumn{4}{l}{\textbf{Fully-Supervised}}\\
\multirow{3}{*}{Fully-Supervised}
% \midrule
& \cellcolor{gray!10} Decision Tree & \cellcolor{gray!10} 81.30 &  \cellcolor{gray!10} 76.97 &  \cellcolor{gray!10} 84.31 & \cellcolor{gray!10} 76.20 & \cellcolor{gray!10} & \cellcolor{gray!10} 80.30 & \cellcolor{gray!10} 53.87 & \cellcolor{gray!10} 88.27 & \cellcolor{gray!10} 52.15 \\
& \cellcolor{gray!10} Logistic Regression & \cellcolor{gray!10} 79.70 & \cellcolor{gray!10} 70.48 & \cellcolor{gray!10} 83.56 & \cellcolor{gray!10} 73.18 & \cellcolor{gray!10} & \cellcolor{gray!10} 80.90 & \cellcolor{gray!10} 58.34 & \cellcolor{gray!10} 86.15 & \cellcolor{gray!10} 59.74 \\
& \cellcolor{gray!10} Random Forest & \cellcolor{gray!10} 78.60 & \cellcolor{gray!10} 66.12 & \cellcolor{gray!10} 83.16 & \cellcolor{gray!10} 70.58 & \cellcolor{gray!10} & \cellcolor{gray!10} 80.20 & \cellcolor{gray!10} 56.49 & \cellcolor{gray!10} 86.14 & \cellcolor{gray!10} 57.34 \\
\midrule
% \multicolumn{4}{l}{\textbf{Machine Learning Methods}}\\
% \midrule
\multirow{3}{*}{Few-Shot (N=6)}
& Decision Tree & 71.10	& 53.14 & 77.62 & 51.16 && 31.90 & 54.81 & 25.99 & 31.71 \\
& Logistic Regression& 58.70 & 73.40 & 53.44 & 56.78 && 53.30 & 53.95 & 53.13 & 48.16 \\
& Random Forest & 69.70 & 62.88 & 72.18 & 63.61 && 65.00 & 51.50 & 68.43 & 51.04 \\
\midrule
% \multicolumn{4}{l}{\textbf{GPT-4}}\\
% \midrule
\multirow{4}{*}{GPT-4}
& Zero-Shot & 51.90 & 76.15 & 42.56 & 51.89 && 24.10 & 51.81 & 16.82 & 22.33 \\
& Zero-Shot+ & 62.90 & 59.30 & 64.29 & 58.58 && 30.00 & 53.25 & 23.76 & 29.67 \\
& Few-Shot (N=6) & 65.70 & 79.35 & 59.89 & 64.72 && 41.20 & 59.05 & 36.33 & 40.88 \\
& \cellcolor{green!10} \ours & \cellcolor{green!10} 79.10 & \cellcolor{green!10} 73.11 & \cellcolor{green!10} 81.43 & \cellcolor{green!10} 73.88 & \cellcolor{green!10} & \cellcolor{green!10} 70.00 & \cellcolor{green!10} 62.88 & \cellcolor{green!10} 71.72 & \cellcolor{green!10} 60.21 \\
\midrule
% \multicolumn{4}{l}{\textbf{GPT-3.5}}\\
% \midrule
\multirow{4}{*}{GPT-3.5}
& Zero-Shot & 78.00 & 66.87 & 82.37 & 68.56 && 56.50 & 59.88 & 55.45 & 52.29 \\
& Zero-Shot+ & 72.40 & 50.00 & 80.37 & 42.00 && 62.60 & 57.62 & 63.96 & 54.40 \\
& Few-Shot (N=6) & 76.30 & 63.73 & 80.93 & 63.84 && 40.80 & 54.56 & 36.96 & 40.32 \\
& \cellcolor{green!10} \ours & \cellcolor{green!10} 79.30 & \cellcolor{green!10} 74.49 & \cellcolor{green!10} 80.98 & \cellcolor{green!10} 71.59 & \cellcolor{green!10}  & \cellcolor{green!10} 66.60  & \cellcolor{green!10} 58.31  & \cellcolor{green!10} 68.83 & \cellcolor{green!10} 55.83 \\
\midrule
% {\ours} & \textbf{79.07 ± 0.31}* & \textbf{82.19 ± 0.13}* & \textbf{71.08 ± 0.17}* & \textbf{41.51 ± 0.25}* & \textbf{79.76 ± 0.18}* & \textbf{70.07 ± 0.13}* & \textbf{40.92 ± 0.12}* & \textbf{61.23 ± 0.18}*\\
\bottomrule
\end{tabular}
}
\vspace{-0.5em}
% We use * to indicate statistically significant results ($p<0.05$) under the same setting.
% }
\vspace{-1em}
\label{tab:main_result}

\end{center}
% \end{adjustbox}
\end{table*}
Table~\ref{tab:main_result} presents the experimental results on the two datasets. The findings highlight several key observations:

\noindent $\diamond$
Traditional machine learning (ML) models achieve respectable performance when fully trained on large datasets (11,353 samples for MIMIC-III and 34,404 samples for CRADLE). However, the performance of simpler models, such as Decision Trees and Logistic Regression, substantially deteriorates in the few-shot learning setting, emphasizing their limitations when labeled data is scarce.

\vspace{-0.5ex}

\noindent $\diamond$
% When comparing the zero-shot or few-shot LLMs with ML methods under the few-shot settings, we find that LLMs tend to demonstrate higher sensitivity but lower specificity. This indicates that LLMs are good at correctly identifying positive cases (i.e., patients with the condition of interest) but at the cost of a higher false positive rate. In other words, LLMs are more likely to classify a patient as having the condition even if they don't. This suggests that LLMs, especially GPT-4, are more conservative, possibly due to the alignment of LLMs to predict false positives to avoid the risk of potentially missing some true positive cases.
When comparing the performance of zero-shot or few-shot LLMs with ML methods under few-shot settings, we observe that LLMs exhibit higher sensitivity but lower specificity. This finding suggests that LLMs excel at correctly identifying positive cases (i.e., patients with the condition of interest) but at the cost of a higher false positive rate. In other words, LLMs are more prone to classifying a patient as having the condition, even when they do not. This tendency implies that LLMs, particularly GPT-4, adopt a more conservative mindset, possibly due to their alignment to err on the side of caution to mitigate the risk of potentially missing true positive cases.

\vspace{-0.5ex}

\noindent $\diamond$
Zero-shot with additional prompting strategies (Zero-Shot+) can improve based on pure zero-shot, with occasionally produced errors. This observation underscores the importance of carefully crafting prompts to optimize the performance of LLMs in EHR-based disease prediction tasks.

\vspace{-0.5ex}

\noindent $\diamond$
Most of the time, adding few-shot demonstrations enhance prediction performance compared to their respective Zero-Shot+ counterparts. This finding emphasizes providing even a limited number of labeled examples can potentially steer language models toward more precise predictions. By leveraging a small set of representative samples, LLMs can quickly adapt to the specific characteristics of the EHR-based disease prediction task. 

\vspace{-0.5ex}

\noindent $\diamond$
Our proposed approach \ours demonstrates remarkable performance, surpassing other methods and even fully supervised ML models in certain scenarios, with GPT-4 generally outperforming GPT-3.5. On the CRADLE dataset, \ours achieves an F1 score of 60.21\%, outperforming all fully trained ML models. Similarly, on the MIMIC-III dataset, \ours obtains an F1 score of 73.88\%, comparable to the fully trained Decision Tree model and superior to Logistic Regression and Random Forest. 
%These results underscore the effectiveness of leveraging collaborative LLM agents for EHR-based disease prediction, particularly when labeled data is limited. 

\vspace{-0.5ex}

\noindent $\diamond$
Compared with the few-shot setting with a single LLM predictor, \ours improves significantly on all four metrics. This can be attributed to the feedback instructions provided by the critic agent, which analyzes the outputs and identifies issues and biases in LLM's reasoning process, such as overly relying on conservative thinking or neglecting certain key factors. The feedback instructions generated by the critic agent help to correct these issues, dynamically refining the predictor agent's reasoning process, thus improving the accuracy of the prediction.

\section*{Generated Instructions}
\vspace{-1em}
\begin{figure}[t]
\centering
  \includegraphics[width=\linewidth]{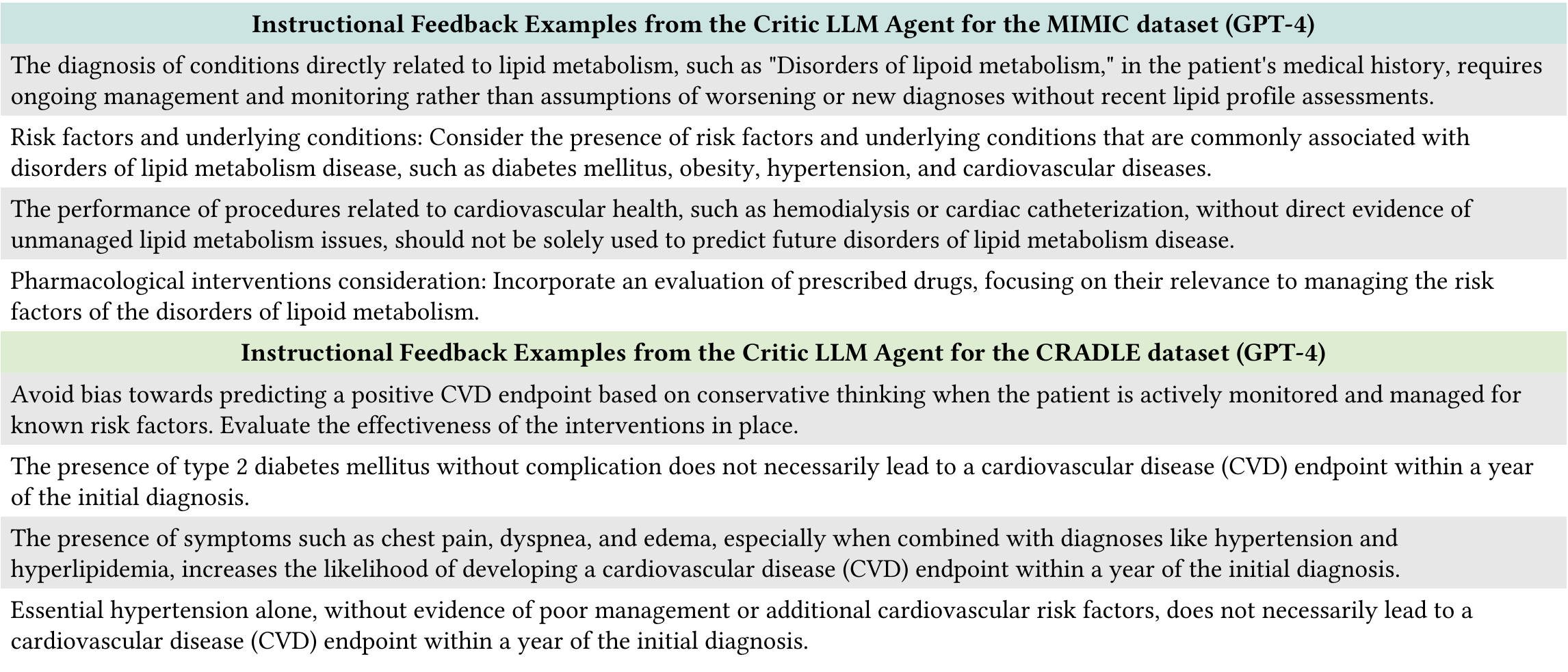}%
  \vspace{-1ex}
    \caption{Examples of instructional feedback generated by the GPT-4-based critic agent, which aims to refine the predictor agent's reasoning process and improve the accuracy of its prediction.}
    \vspace{-2ex}
    \label{fig:criteria}
    % \vspace{-1em}
\end{figure}

Figure~\ref{fig:criteria} showcases examples of the criteria and instructions generated by the critic agent. These examples demonstrate the critic agent's ability to identify potential issues in the predictor agent's prediction and reasoning process and provide targeted instructions to address them. For instance, the first instruction for the CRADLE dataset, ``Avoid bias towards predicting a positive CVD endpoint based on conservative thinking when the patient is actively monitored and managed for known risk factors. Evaluate the effectiveness of the interventions in place" highlights a possible prediction bias of the predictor agent. This instruction encourages the predictor agent to avoid relying on conservative assumptions when making predictions, as such assumptions may be a result of the over-alignment of advanced AI models. By explicitly addressing this issue, the critic agent aims to guide the predictor agent toward more objective and comprehensive reasoning.
Another example for the MIMIC dataset, ``Pharmacological Interventions Consideration: Incorporate an evaluation of prescribed drugs, focusing on their relevance to managing the risk factors of the disorders of lipoid metabolism" suggests that the predictor agent should take into account the role of prescribed medications in managing the patient's condition. By analyzing the relevance and potential impact of these drugs on the risk factors associated with disorders of lipoid metabolism, the predictor agent can make more informed predictions. 
These examples illustrate how the critic agent's feedback can guide the predictor agent towards more comprehensive and nuanced reasoning, ultimately leading to improved disease prediction performance.

\vspace{-0.5em}
\section*{Conclusions}
\vspace{-1em}
% In this study, we investigated the feasibility of applying Large Language Models (LLMs) to Electronic Health Record (EHR) based disease prediction tasks. By converting structured patient visit data into natural language narratives, we evaluated the zero-shot and few-shot diagnostic performance of LLMs using various prompting strategies. Furthermore, we proposed a novel approach that combines two LLM agents with different roles: a predictor agent that generates predictions and reasoning processes, and a critic agent that analyzes incorrect predictions and provides guidance for improving the predictor agent's performance. This collaborative approach enables the system to learn from its mistakes and adapt to the specific challenges of EHR-based disease prediction. Our work highlights the potential of LLMs as a tool for clinical decision support and contributes to the development of efficient and effective disease prediction systems that can operate with minimal training data. 
In this study, we investigated the application of Large Language Models (LLMs) to Electronic Health Record (EHR) based disease prediction tasks. We evaluated the zero-shot and few-shot diagnostic performance of LLMs using various prompting strategies and proposed a novel collaborative approach combining a predictor agent and a critic agent. This approach enables the system to learn from its mistakes and adapt to the challenges of EHR-based disease prediction. Our work highlights the potential of LLMs as a tool for clinical decision support and contributes to the development of efficient disease prediction systems that can operate with minimal training data.

% \section*{Acknowledgements}
% \vspace{-1em}
% This research was partially supported by the internal funds and servers provided by the Computer Science Department of Emory University.
% MKA was partially supported by the Georgia Center for Diabetes Translation Research, funded by the National Institute of Diabetes Digestive and Kidney Disorders (P30DK111024).
% JCH was supported by NSF grants IIS-1838200, IIS-2145411 and NIH grant 5K01LM012924-03.
\vspace{-0.5em}
\section*{Ethical Considerations}
\vspace{-1em}
To ensure the ethical use of credential data with GPT-based services, we have signed and strictly adhered to the PhysioNet Credentialed Data Use Agreement\footnote{https://physionet.org/about/licenses/physionet-credentialed-health-data-license-150}. We follow the guidelines\footnote{https://physionet.org/news/post/gpt-responsible-use} for responsible use of MIMIC data in online services, including opting out of human review of the data through the Azure OpenAI Additional Use Case Form\footnote{https://aka.ms/oai/additionalusecase}, to prevent sensitive information from being shared with third parties.

% References as numbers
\makeatletter
\renewcommand{\@biblabel}[1]{\hfill #1.}
\makeatother

% \newpage 
% \clearpage
% unstr is used to keep citation order
\bibliographystyle{vancouver}
\bibliography{amia}  

% \section*{Appendix}
% \vspace{-1em}
% \input{prompt_predict_agent}
% \input{prompt_critic_agent}

\end{document}